\pdfoutput=1
\documentclass[11pt]{article}

\usepackage{acl}

\usepackage{times}
\usepackage{latexsym}

\usepackage[T1]{fontenc}
\usepackage[utf8]{inputenc}
\usepackage{times}
\usepackage{latexsym}

\usepackage{enumitem}
\usepackage{amsmath}
\usepackage{commath}

\usepackage{graphicx}
\usepackage{longtable}
\usepackage{multirow}
\usepackage{multicol}
\usepackage{float}
\usepackage{colortbl}
\usepackage{microtype}

\title{A Survey of Parameters to Assess the Quality of Benchmarks in NLP}
\title{A Survey of Parameters Associated with the Quality of Benchmarks in NLP}

\author{Swaroop Mishra $\;$ Anjana Arunkumar $\;$ Chris Bryan $\;$ Chitta Baral
\\\\
 Arizona State University }

\begin{document}
\maketitle
\begin{abstract}
Several benchmarks have been built with heavy investment in resources to track our progress in NLP. Thousands of papers published in response to those benchmarks have competed to top leaderboards, with models often surpassing human performance. However, recent studies have shown that models triumph over several popular benchmarks just by overfitting on spurious biases, without truly learning the desired task. Despite this finding, benchmarking, while trying to tackle bias, still relies on workarounds, which do not fully utilize the resources invested in benchmark creation, due to the discarding of low quality data, and cover limited sets of bias. A potential solution to these issues- a metric quantifying quality- remains underexplored. Inspired by successful quality indices in several domains such as power, food, and water, we take the first step towards a metric by identifying certain language properties that can represent various possible interactions leading to biases in a benchmark. We look for bias related parameters which can potentially help pave our way towards the metric. We survey existing works and identify parameters capturing various properties of bias, their origins, types and impact on performance, generalization, and robustness. Our analysis spans over  datasets and a hierarchy of tasks ranging from NLI to Summarization, ensuring that our parameters are generic and are not overfitted towards a specific task or dataset. We also develop certain parameters in this process. 
% Our survey identifies gaps in the current research and suggests potential future research directions such as a metric quantifying quality, demonstrating opportunities for the next generation of benchmarks.
\end{abstract}

\section{Introduction}

Large-scale Benchmarks such as SNLI \cite{bowman2015large}, SQUAD \cite{rajpurkar2016squad}, GLUE \cite{wang2018glue}, and Senteval \cite{conneau2018senteval} have been guiding our progress in NLP over the years. Benchmark creation and subsequent model development have involved heavy investment in resources such as money and time. 
%Models devised in response to these benchmarks in turn compete to top leader boards, with high consumption of several resources, including computation. 
Often, language models such as BERT \cite{devlin2018bert}, RoBERTA \cite{liu2019roberta} and GPT3 \cite{brown2020language} beat human performance. A growing number of recent works \cite{gururangan2018annotation, poliak2018hypothesis, kaushik2018much, tsuchiya2018performance, tan2019investigating, schwartz2017effect, swayamdipta-etal-2020-dataset, gardner-etal-2021-competency, pezeshkpour2021combining} however, expose an undesired reason this can happen: instead of learning tasks like humans, models simply overfit to spurious biases. Therefore, we must reexamine the process of creating and solving benchmarks.

Several algorithms \cite{sakaguchi2019winogrande,li2019repair,li2018resound, wang2018dataset,clark2019don,he2019unlearn,mahabadi2019simple,zellers2018swag,nie2019adversarial,kaushik2019learning,gardner2020evaluating} have been proposed over the past few years to tackle the issue of bias\footnote{Bias represents spurious bias unless otherwise stated.}. However, there is significant opportunities as we need methods that can (i) provide a metric to quantify quality, (ii) justify the original investment in benchmark creation as they discard bad samples, (iii) cover diverse categories of bias, (iv) are generic, i.e. independent of models and tasks, and automated, i.e., free of human intervention in judging data quality. A metric that quantifies benchmark quality could serve as a single potential solution to all these issues. An important question which has remained underexplored over the years is: \textit{how can we assess and quantify the quality of benchmarks?}

In pursuit of this goal, we draw inspiration from successful quality indices in various domains such as power \cite{bollen2000understanding}, water \cite{world1993guidelines}, food \cite{grunert2005food} and air \cite{jones1999indoor}.

\begin{figure*}
\includegraphics[width=\linewidth]{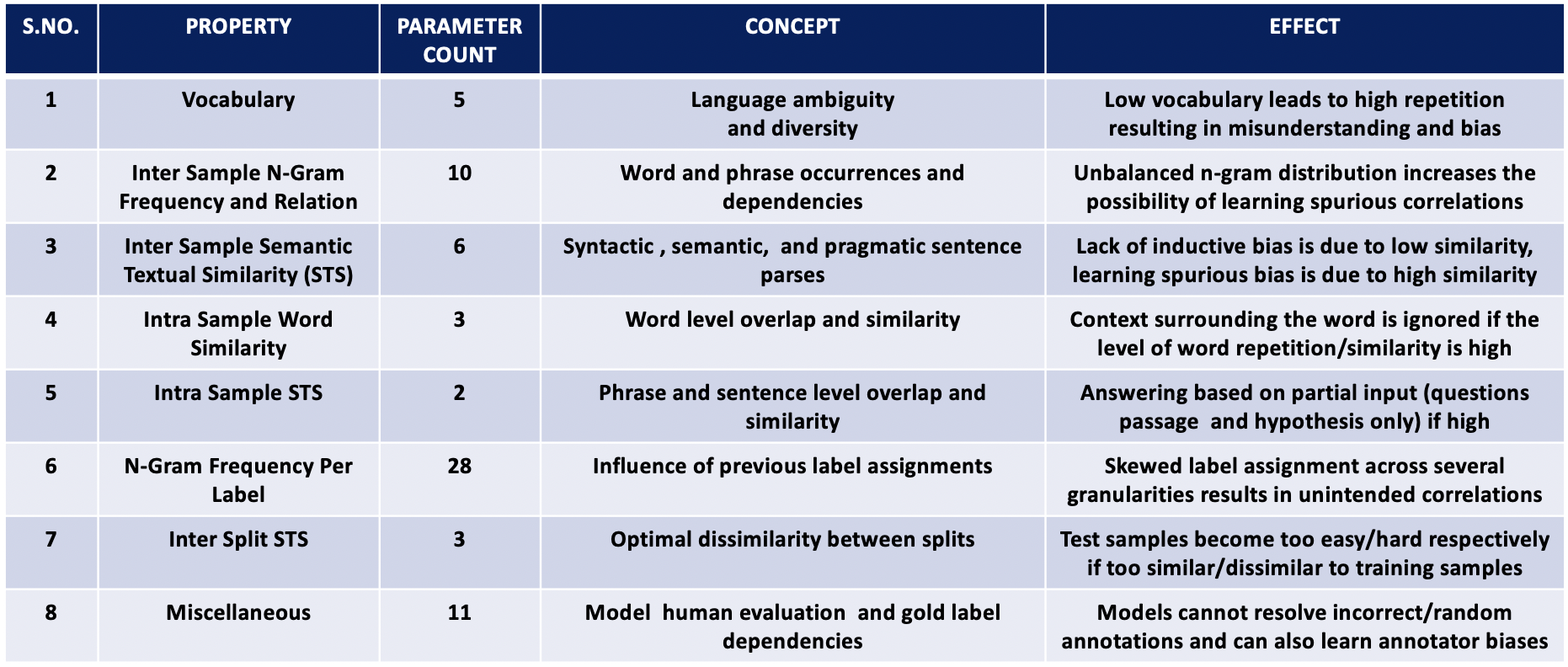}
  \caption{Column 2: list of language properties we have identified, Column 3: count of associated parameters, Column 4: underlying concepts, and Column 5: potential effects of properties on bias.}
\label{fig:seven}
% \vspace{-2mm}
\end{figure*}
This paper focuses on the initial steps to develop a metric quantifying data quality for NLP benchmarks. In our study, higher data quality implies lower bias and higher generalization capability. To construct such a metric, we first identify a set of seven language properties to represent various possible sample interactions (and therefore biases) in a benchmark. For each property we identify potential bias parameters that can be integrated into the metric. This is done by surveying existing literature, to understand how various parameters capture bias properties, including origins, types and impact on performance, generalization, and robustness. We also develop several new parameters in this process. To ensure that identified parameters are generic and not overfitted, we cover several datasets and a hierarchy of tasks: NLI, Argumentation, QA, Reading Comprehension (RC), and Abstractive Summarization. This order reflects the presence of increasing amounts of data per sample across tasks. 

% By identifying salient bias parameters and categorizing them according to their language properties, we lay the foundation for the development of a novel quality metric for NLP benchmarks. 
% In a concurrent work, Anonymous et.al.~[1]\footnote{Anonymous reference content can be found in the supplemental material} uses our findings directly to develop a metric quantifying benchmark quality. Anonymous et.al.~[2] uses the metric to build a benchmark creation platform that actively educates and discourages users from creating biased data samples. Anonymous et.al.~[3] further develops a paradigm to assist crowd-workers in handling bad quality data with the help of user-recommendations and adversarial transformation.

\section{Components of the Metric}
\label{sec:ttwwoo}
\begin{figure}[H]
\includegraphics[width=7cm]{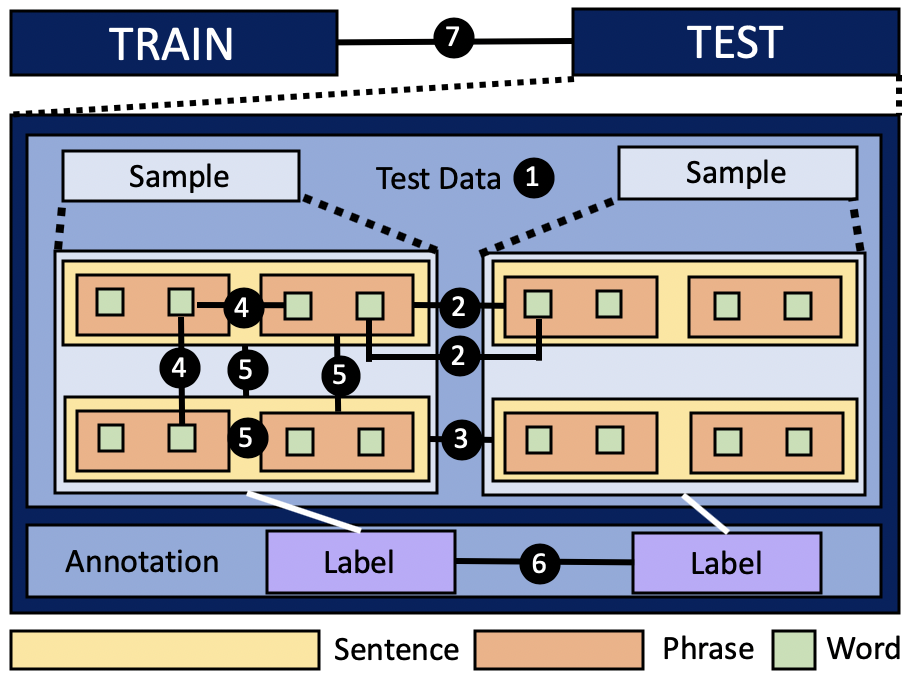}
  \centering
  \caption{Hierarchy used in defining properties to exhaustively cover all possible sample interactions}
\label{fig:model}
\vspace{-1mm}
\end{figure}

We identify seven text properties- as illustrated in Figure~\ref{fig:seven}- that cover various possible inter/intra-sample interactions (a subset of which leads to biases) in an NLP dataset. The properties are intuitively selected by considering interaction hierarchies present at word, phrase, sentence, data, annotation, and split levels, as illustrated in Figure~\ref{fig:model}. These can be used to define metric components.

\section{Literature Survey to Identify Bias Parameters}
We survey 77 papers in three stages: (i) identifying and compiling diverse parameters\footnote{parameters marked with * are new, the rest are either directly compiled or modified from existing works} related to bias impact, identification, isolation, and removal in various datasets, (ii) extrapolating parameters developed for a particular NLP task to a broader task set, and (iii) curating parameters by relating any model failures to potential bias. The parameters are binned into seven properties representing various components of the metric as discussed in Section \ref{sec:ttwwoo} and listed in Figure~\ref{fig:seven}.

\section{Survey Results- Potential Parameters}\label{plead}
In this section, we comprehensively list 68 potential parameters that either (i) directly indicate bias, (ii) inspect possible bias existence via model probing, or (iii) can be utilized to remove bias.  We consider a range of NLP tasks, in the following order: NLI, Argumentation, QA, Reading Comprehension, and Abstractive Summarization (as per increasing amounts of data per sample across tasks)\footnote{Examples, and more details in the supplemental materials.}. We follow this order while looking for parameters as bias analysis on lower order tasks can be extended to higher order tasks. This helps further in developing a generic metric from these parameters.
% \vspace{-2mm}
\subsection{Vocabulary}
\paragraph{Vocabulary Magnitude:*}
We define this as the ratio of dataset vocabulary size to the size of the dataset. The performance drop for MNLI is lesser than SNLI on providing partial input. Lower drop has been attributed to the presence of multiple genres in MNLI \cite{poliak2018hypothesis,gururangan2018annotation}.  This may indicate that high vocabulary magnitude is desirable, and will reduce model dependency on spurious correlations. 
% \vspace{-2mm}
\paragraph{Vocabulary across POS Tags:*}
Vocabulary needs to be examined across POS tags to account for the presence of homonyms in vocabulary. The word distribution across samples might also be a useful bias indicator.
% \vspace{-2mm}
\paragraph{Language Perturbation:}
Correlations exploited by models can be exposed by isolating cases in which certain words or phrases are not used as a part of context in answering. Isolation can be achieved through the generation of examples by replacement of conjunctive \cite{talmor2019olmpics} phrases with meaningless filler words. If the model has low language sensitivity, it will not show any change in the learning curve for these perturbations. This can also be used to evaluate the influence of prepositional phrases, and thus the associated bias.
% \vspace{-7mm}
\paragraph{Semantic Adverb Resolution:}
The ability of models to correctly perceive  and differentiate the usage of adverbs such as ‘always’, ‘sometimes’, ‘often’, and ‘never’ reflects the extent of its reasoning capabilities \cite{talmor2019olmpics}. The relationship between the model performance and level of presence of adverbs across samples is a viable parameter to investigate bias.
% \vspace{-2mm}
\paragraph{Domain Specific Vocabulary:*}
The presence of multiple genres in datasets dilutes bias influence \cite{poliak2018hypothesis,gururangan2018annotation,glockner2018breaking}, and these datasets have a large amount of domain specific vocabulary (e.g.: ordinals, nationalities, countries, etc.). Therefore, the presence of an increased number of domain specific words seems desirable to minimize artifacts.
\subsection{Inter-sample N-gram Frequency and Relation}
\paragraph{Maximal Word Distance:*}
The presence of multiple genres accounts for the robustness of the MNLI dataset in comparison to SNLI \cite{poliak2018hypothesis,gururangan2018annotation}. This can be quantified, in terms of spreading the ‘distances’ between word vectors to the maximum extent. Higher robustness might indicate lesser bias.
\paragraph{POS Tag Replacement:}
POS tag replacement is a method to increase the vocabulary size in a controlled manner, and balance the word distribution of a dataset. 
Erasure \cite{li2016understanding} sometimes generates semantically or grammatically incorrect sentences \cite{zhao2017generating, jin2019bert}. Occurrence of incorrect sentences can be reduced by replacing sentence tokens with random words of the same POS tag probablistically, based on their embedding similarities \cite{ribeiro-etal-2018-local}. Combining POS tag replacement with the approach of discarding sentences that contain bigrams of low frequencies \cite{glockner2018breaking} can further improve this method.  POS Tag Replacement may help to reduce the bias content in a dataset.
%Similarly, Textfooler \cite{jin2019bert} is a recent work which creates adversarial data by replacing words.
\paragraph{Consecutive Verb Frequency:}
Machine translation results in dropping of consecutive verbs \cite{zhao2017generating}. We extrapolate this as a bigram related parameter for estimating bias content. 
% \vspace{-2mm}
\paragraph{Anonymization of Entities:}
Masking of entities can help in understanding the reliance of models on co-occurrence based biases while attaching a role to that entity. This has been used in the cloze style preparation of samples in RC datasets \cite{hermann2015teaching}. Similar type of representation bias is also addressed in Vision \cite{li2018resound}, in terms of object, scene and person bias.
% \vspace{-2mm}
\paragraph{Metonymy:}
The usage of figures of speech in sentences will help in identifying whether models are using context effectively \cite{clark2018knowledge}. It further provides a case to examine model dependency on word association.
% \vspace{-2mm}
\paragraph{Stereotypes:}
Hypotheses in NLI datasets contain gender, religious, race and age based stereotypes \cite{rudinger2017social}. This can be a form of contextual bias, in that the occurrence of sets of stereotype n-grams could bias the model towards a particular label. 
% \vspace{-2mm}
\paragraph{Out of Distributions in Range}
Models relying on spurious correlations fail on out of sample distributions~\cite{bras2020adversarial, mishra2020our, gokhale2022generalized}.  For example, RoBERTA cannot resolve numbers to be ages if they are not in a typical human range \cite{talmor2019olmpics}. Evaluation with just in-domain samples are shown to be not robust in reflecting true capability of models~\cite{mishra2021robust}. Evaluation of out of distribution tasks i.e. cross-task generalization~\cite{mishra2022cross, parmar2022boxbart} have recently been popular owing to the sucess of large language models~\cite{brown2020language}.
% \vspace{-2mm}
\paragraph{Handling Conjunctions:}
Models have difficulty in determining if conjunctional clauses are true, which is necessary in reasoning inference chains based on sorting, and comparison \cite{talmor2019olmpics}. Handling  conjunctions effectively could eliminate model reliance on bias to solve reasoning based tasks.
% \vspace{-2mm}
\paragraph{Unnatural Language:}
This refers to contradictory phrase pairs that arise by substituting adjectives and adverbs of opposing intent. For example, the usage patterns of ‘not’ and ‘very’ are identical in some cases, though the sentence meanings are opposite \cite{talmor2019olmpics}. This can be a useful probe to evaluate dependency of models on bias. 
% \vspace{-2mm}
\label{sec:49}\paragraph{Broad Referring Expressions:}\hyperref[sec:eg49]{(e.g.)}
The use of ‘broad’ referring expressions  like ‘the’, ‘this’, ‘that’, and ‘it’ in a test set distribution serves to test the ability of a model to reason based on any referential resolution patterns it has identified in the training set \cite{gundel1993cognitive,mcshane2016resolving,degen2020redundancy}.

\subsection{Inter-sample STS}
\paragraph{Sentence Structure:}
If the distribution of parse structures is skewed, i.e., a small proportion of parse trees dominates the majority of training samples, models may just learn spurious correlations, and thus perform poorly \cite{poliak2018hypothesis}. 
% \vspace{-2mm}
\paragraph{Multistep Reasoning:}
Failure to do multistep reasoning might be an indicator of learning spurious correlations. This is evinced by two cases, compositional \cite{talmor2019olmpics} and numerical reasoning. \cite{naik2018stress}.
Both patterns follow a chain of inferences, with numerical reasoning additionally quantifying and solving arithmetic questions.
% \vspace{-2mm}
\paragraph{Inter-Sentence Antithesis:}
A special case of pattern exploitation in language modelling  is in converse examples, wherein two samples have identical linguistic patterns, and only differ with a single word or phrase of opposing meaning \cite{naik2018stress}. Incorrect resolution of this case might suggest a model's dependency on annotation artifacts.
% \vspace{-7mm}
\paragraph{Sentence Length Variation:}
Sentence length should vary across samples to prohibit models from using it as an artifact. \cite{gururangan2018annotation}.
% \vspace{-2mm}
\paragraph{Start Tokens:}
The presence of repeated start tokens in the premise and hypothesis, could bias a model to focus only on certain parts of the input, thus it is an useful lead. \cite{sugawara2018makes}. 
% \vspace{-2mm}
\paragraph{Ellipsis Resolution:}
The presence of ellipsis in samples has been a point of shortfall for language models \cite{clark2018knowledge}, due to their reliance on factitious relations. So, this can be an effective probe.
\subsection{Intra-sample Word Similarity}
\paragraph{Presupposition and Query:}
Sometimes, sentences indicate an already implied fact, which is utilized for a further query on a specific case of that fact \cite{clark2018knowledge}. 
% \vspace{-2mm}
\paragraph{Coreference Resolution:}
Coreferences can be a result of the usage of pronouns, as well as abstractive words like ‘each’ and ‘some’ \cite{gururangan2018annotation,cirik2018visual} and can occur within or between statements. The inability to correctly resolve coreferences may suggest the misunderstanding of context, which has possibly led to bias.
% \vspace{-7mm}
\paragraph{Taxonomy Trees:}
Consider a conjunctive phrase with two objects that can be grouped under a generic super-class. Because of the effect of bias, the first object’s closest parent on the taxonomy tree is often taken as the superset across both objects. For example, ‘horse and crow’ would be grouped as ‘animal’, but ‘crow and horse’ may be grouped as ‘bird’ in some cases \cite{talmor2019olmpics}. 
\subsection{Intra-sample STS}
\paragraph{Overlap:}
Overlap in words among premise-hypothesis pairs could indicate label.
Failure to resolve antonymy and negation is a special case of this \cite{naik2018stress}. This bias indicating feature is used in the construction of the adversarial dataset HANS, in three ways: (i) assuming a premise entails all hypotheses constructed from premise words, (ii)assuming a premise entails all its contiguous subsequences, and (iii) assuming a premise entails all complete subtrees in its parse tree.
\cite{mccoy2019right}. A similar measure with division of total number of words as a normalization term  has also been proposed in another work \cite{dasgupta2018evaluating}.
% \vspace{-2mm}
\paragraph{Sentence Similarity:}
High sentence similarity among premise-hypothesis pair biases systems towards assigning the label of ‘entailment’, and low similarity towards ‘neutral’ \cite{naik2018stress}. 
\subsection{N-gram Frequency per Label}
\paragraph{Erasure:}
Different levels of representations used by models are erased to find the change in model output. Several techniques such as reinforcement learning \cite{li2016understanding} has been used to erase minimal sets of input words and flip model decisions.
This technique can indirectly help identify elements producing artifacts by extrapolating the minimal set of input words responsive to models.
% \vspace{-2mm}
\paragraph{Negation:}
Samples containing universal negation terms such as ‘no’, ‘not’ are predisposed to the contradiction label. \cite{poliak2018hypothesis}.
% \vspace{-2mm}
\paragraph{Antonymy:}
Discarding antonymy due to the absence of explicit negation indicates model bias \cite{naik2018stress}. 
% \vspace{-2mm}
\paragraph{WL Mapping:}
This parameter measures the level of correlation within a class label. $P(l/w)$ is the conditional probability of the occurrence of a label $l$ given a word $w$. If it has value 0 or 1, the label becomes trivial \cite{poliak2018hypothesis}. Such a skew leads to inference on the basis of word presence, a spurious bias.
% \vspace{-2mm}
\paragraph{PL Mapping:}
Pattern exploitation can be extended to phrase level dependencies of labels, measured as $P(l/p)$, i.e. $P(label/phrase)$ \cite{poliak2018hypothesis}. 
% \vspace{-2mm}
\paragraph{Vocabulary Score:*}
We define this as a constant length vector of: (i) the number of labels a given word is present in, (ii) the individual counts of the word in each label. 
This parameter will help estimate and prevent the skewness of a word distribution towards a label;
for example, the word `sleep’ and its variations are found to be indicators of contradiction, as they were predominantly present in samples with that label \cite{poliak2018hypothesis}.
% \vspace{-2mm}
\paragraph{Copying:}
Copy augmented modeling has been used in the split and rephrase task \cite{aharoni-goldberg-2018-split,gu-etal-2016-incorporating}, abstractive summarization \cite{see-etal-2017-get}, and language modelling \cite{merity2016pointer}. We extend this parameter and propose the use of an iterative copy mechanism, to copy different word n-grams between the premise and hypothesis in NLI. By noting the points of the label change, we can isolate the most informative word overlap sets.
% \vspace{-2mm}
\paragraph{Hypothesis Only Prediction:}
This parameter is used to test dependencies between the label and hypothesis, to prevent partial answering based on spurious correlation \cite{tan2019investigating}.
% \vspace{-2mm}
\paragraph{Cue Influence:}
Nature of artifacts and their contribution to Warrant only predictions has been studied for the ARCT dataset \cite{niven2019probing}. Their evaluation using three metrics: applicability, productivity, and coverage can be extended to other tasks for finding the influence of cues.
% \vspace{-2mm}
\paragraph{Length Mismatch:}
The length of a sentence can indicate its label class, as entailment or neutral for shorter and longer sentences respectively. Additionally, length mismatches between the premise and hypothesis can predispose the model to predict non-entailment labels \cite{poliak2018hypothesis,gururangan2018annotation,naik-etal-2018-stress}.
% \vspace{-2mm}
\paragraph{Grammaticality:}
Tests on the FN+ dataset show that sentences with poor grammar are classified under non-entailment labels \cite{poliak2018hypothesis}.
% \vspace{-2mm}
\paragraph{PMI:}
PMI is a popular parameter that represents the word-label dependency. It measures how likely a word and a label are to co-occur, given their independent probabilities, and joint probability  \cite{naik2019exploring,gururangan2018annotation}.
% \vspace{-2mm}
\paragraph{Scripts:}
A way to break down complex inference chains is to identify common scripts based on the incorporation of real world knowledge \cite{clark2018knowledge} . For example, ‘X wants power and therefore tries to acquire it, Y doesn’t want X to have power and tries to thwart X’ is a common script for inference chains. A biased model will fail to identify scripts.
% \vspace{-2mm}
\paragraph{Numerical Reasoning:}
The accurate understanding of numbers is essential to predict the correct label. Language models often fail at numerical reasoning \cite{naik2018stress, wallace2019nlp, mishra2022numglue, lu2022learn}. Also, the presence of numbers predisposes bias to entailment, as entailment examples are seen to have numerical information abstracted with words like ‘some’ or ‘few’ \cite{gururangan2018annotation}.
% \vspace{-2mm}
\paragraph{Gender:}
The absence of gender information indicates entailment \cite{gururangan2018annotation}.
% \vspace{-2mm}
\paragraph{Hypernyms and Hyponyms:}
Models follow a super-set/sub-set structured approach, in the form of hypernyms and hyponyms \cite{richardson2019does}, while assigning the entailment label. Entailment samples are generated by replacing words with their synonyms, hyponyms and hypernyms \cite{glockner2018breaking}. Contradiction samples are generated by replacing words with mutually exclusive co-hyponyms and antonyms. Co-hyponym resolution is an issue for biased models. Therefore, the above methods of sample generation produce adversarial samples . Models using DIRT \cite{lin2001dirt} based methods suffer from the problem of forming prototypical hypernyms as spurious biases while solving. For example, a chair might serve as a super-set for its legs, even though it is not a true hypernym \cite{levy2015supervised}.
% \vspace{-2mm}
\paragraph{Modifiers and Superlatives:}
The use of modifiers such as `tall' and `sad', and superlatives like `first’ and `most’ is predominantly seen in the neutral class \cite{gururangan2018annotation}.
% \vspace{-2mm}
\paragraph{Causal Phrases:}
Phrases like `because of’ and `due to’ are associated with the neutral class, as they add specificity \cite{gururangan2018annotation}.
% \vspace{-2mm}
\paragraph{Absence Indicators:}
Words like ‘sleep’ or ‘naked’ indicate the absence of an object in the sentence, and therefore are associated primarily with the contradiction class \cite{gururangan2018annotation}.
% \vspace{-2mm}
\paragraph{Ambiguity:}
Cases where external knowledge or chain reasoning is required to solve referential cues are classified as neutral \cite{naik2018stress}.
%\label{sec:45}\paragraph{Bigram Entropy:}\hyperref[sec:eg45]{(e.g.)}
% \vspace{-2mm}
\paragraph{Bigram Entropy:}
High entropy bigrams can be used as indicators of entailment and neutral labels \cite{tan2019investigating}: 
This can be extended to phrases. Similarly, in Vision, extrapolation can be done on the forms of representation bias \cite{li2018resound}, in the form of object, scene, and person bias.
% \vspace{-2mm}
\paragraph{Paraphrasing:}
Paraphrased question generation is often used to generate additional samples \cite{sugawara2018makes}. PAWS, an adversarial dataset for paraphrase identification, employs word swapping and back translation to generate challenging paraphrase pairs \cite{zhang2019paws}. However, the limit of paraphrasing is an important lead to be considered, i.e., at what point does the semantic meaning change? An example of this is the inability of a model to distinguish between the meanings of ‘same’ and ‘about the same’ \cite{clark2018knowledge}. 
% \vspace{-2mm}
\paragraph{Multiple Cases:}
This parameter deals with the possible ambiguity in answer choice selection \cite{sugawara2018makes}. This occurs when there are multiple span matches among answer choices to the passage span selected by the question. In the context of NLI, this can be viewed as an indicator for neutral and non-neutral label assignment.
% \vspace{-2mm}
\paragraph{Modality and Belief:*}
Modality details how things could, must, or could not have been. Belief is viewed as a true/false construct when deciding if a modality holds for NLI. This is reflected in patterns followed by human annotators \cite{bowman2015large,williams2017broad}. 
% \vspace{-2mm}
\paragraph{Shuffling Premises:}
Shuffling premises and checking model output can indicate the influence of premises in deciding label \cite{tan2019investigating}. 
% \vspace{-2mm}
\paragraph{Concatenative Adversaries:}
The addition of distracting phrases added in conjunction with premise hypothesis pairs might help test the model’s reliance on spurious biases  \cite{naik2018stress,jia2017adversarial}.
% \vspace{-2mm}
\paragraph{Crowdsource Setting:}
Analysis of the story cloze task \cite{mostafazadeh-etal-2016-corpus} shows that there is a difference in the writing styles employed by annotators in different sub-tasks \cite{schwartz2017effect}. 
Following patterns are observed: (i) decrease in sentence length, (ii) fewer pronouns, (iii) decrease in use of coordinations like `and', (iv) less enthusiastic and increasingly negative language. These are also found to be indicators of deceptive text \cite{qin2004exploratory} which have been classified in to five categories: quantity, vocabulary complexity, sentence complexity, specificity and expressiveness, and informality, on the basis of nineteen parameters. The mean number of clauses per utterance and the Stajner- Mitkov measure of complexity have also been included as highly informative syntactic features for deception in text \cite{yancheva2013automatic}. Liars tend to use fewer self-references, more negative emotion words, and fewer markers of cognitive complexity, i.e., fewer `exclusive' words, and more `motion' verbs like walk and go \cite{newman2003lying}. Crowdworkers are also shown to ammplify the patterns present in the instructions given to them~\cite{parmar2022don}.
% \vspace{-2mm}
\paragraph{Sample Perturbation:}
Counterfactual samples are created using a human-in-the-loop system \cite{kaushik2019learning}. When a model is trained on these samples, it fails on the original data, and vice versa. However, augmenting the revised samples reduces correlations formed from each set individually. Recently, contrast sets \cite{gardner2020evaluating} have been created by perturbing samples to change the gold label, to view a model's decision boundary around a local instance. Model performance on contrast sets decreases, creating new benchmarks.
\subsection{Inter-split STS}
\paragraph{Variation of Split:}
Recent studies have shown that benchmarking is done improperly, due to the presence of fixed training and test sets \cite{tan2019investigating}. Also, evaluation metrics are mistakenly treated as exact quantities. They should instead be treated as estimates of random variables corresponding to true system performance. Many works either do not use proper statistical tests- such as hypothesis testing- for system comparison or do not report test results \cite{gorman-bedrick-2019-need}. 
% \vspace{-8mm}
\paragraph{Annotator Bias:}
Model performance improves on including annotator identifiers as training features. Models do not generalize if the annotators of test set have not contributed in annotating the training set, implying that models fit to annotators and not tasks. Keeping annotators across train and test spilt disjoint can mitigate bias
\cite{geva2019we}.
% \vspace{-2mm}
\paragraph{World Definition:}
The negative set of a dataset defines what it considers to be “the rest of the world”. If that set is not representative, or unbalanced, it could produce overconfident and not discriminative classifiers \cite{torralba2011unbiased}. 
\subsection{Miscellaneous}
We also identified several bias parameters dealing with model interaction, human evaluation, and gold-label determination. These cannot be sorted into the previous categories as we desire a data quality metric that is (i) model-independent (ii) not considering any flaws in assigning gold labels.
\paragraph{Inoculation Cost:}
It is defined as the improvement in model performance post inoculation \cite{richardson2019does}. Inoculation involves training on new tasks using small sample sets. This aims at fine tuning the model to perform robustly on out of distribution samples without re-purposing the model entirely. Adversarial NLI \cite{nie2019adversarial} uses an adversarial human-and-model-in-the-loop procedure, to generate a new adversarial dataset, on which a model is trained to improve performance. However, both these approaches may introduce their own set of model-dependent biases. 
% \vspace{-7mm}
\paragraph{Disagreement:}
If disagreement amongst annotators is a random noise, then data with low reliability can be tolerated by model. However, if the disagreement contains patterns, then a model can use these patterns as a spurious bias to boost its performance. By testing for correlation between two annotators, some of these patterns can be identified. However, not all patterns picked up by the model  will necessarily show up on the correlation test- a scenario where the number of samples with disagreement is too low \cite{reidsma2008reliability}.
% \vspace{-2mm}
\paragraph{Random Labelling:}
Surprisingly, models achieve zero training error on datasets where the true labels are replaced by random labels. 
% So, without changing the model, model size, hyper parameters, and optimizer, the generalization error of a model can be forced to increase considerably. 
% Stochastic gradient descent with unchanged hyper parameter settings can optimize weights to fit to random labels perfectly, even though the true meaning of the labels is lost
Explicit regularization techniques like weight decay and dropout are found to be insufficient for controlling generalization error. \cite{zhang2016understanding}.
% \vspace{-2mm}
\paragraph{Re-Optimizing Weights:}
REPAIR formulates bias minimization as an optimization problem, by redistributing weights to penalize easy examples for a classifier \cite{li2019repair}. 
% \vspace{-2mm}
\paragraph{Adversarial Filtering:}
AFLite \cite{sakaguchi2019winogrande} randomly samples data in to train and test set. It uses a linear model to classify, and identifies the most correctly classified examples as the biased examples, which subsequently get removed.
% \vspace{-2mm}
\paragraph{Ranking Artifacts:*}
We propose that artifacts need be ranked based on the extent of their influence on label. Using this ranking, the artifact combinations and occurrences that give rise to a greater amount of bias can be isolated.
% \vspace{-2mm}
\paragraph{Human Performance Measurement:*}
In several works, authors are involved in measuring human performance \cite{gardner2020evaluating} . Since, authors know the intricacies of the dataset creation process, human baseline gets biased.
% \vspace{-2mm}
\paragraph{Order of Input:}
The order in which training data is fed to the model affect its performance \cite{dodge2020fine}. These orderings are particular to a dataset, and thus may provide new leads to bias. 
% \vspace{-2mm}
\paragraph{Models of Annotation:}
Bayesian models of annotation have been shown to be better than traditional approaches of majority voting and coefficients of agreement. This finding is based on the study to improvise the traditional way of calculating and handling gold standard labels, annotator accuracies and bias minimization, item difficulties and error patterns \cite{paun2018comparing}.
% \vspace{-2mm}
\paragraph{Exposure Bias:}
A model's way of handling data may introduce bias such as the exposure bias which is introduced because of the difference in exposing data to the model during training and inference phase \cite{caccia2018language}.
\begin{figure*}
\includegraphics[width=\linewidth]{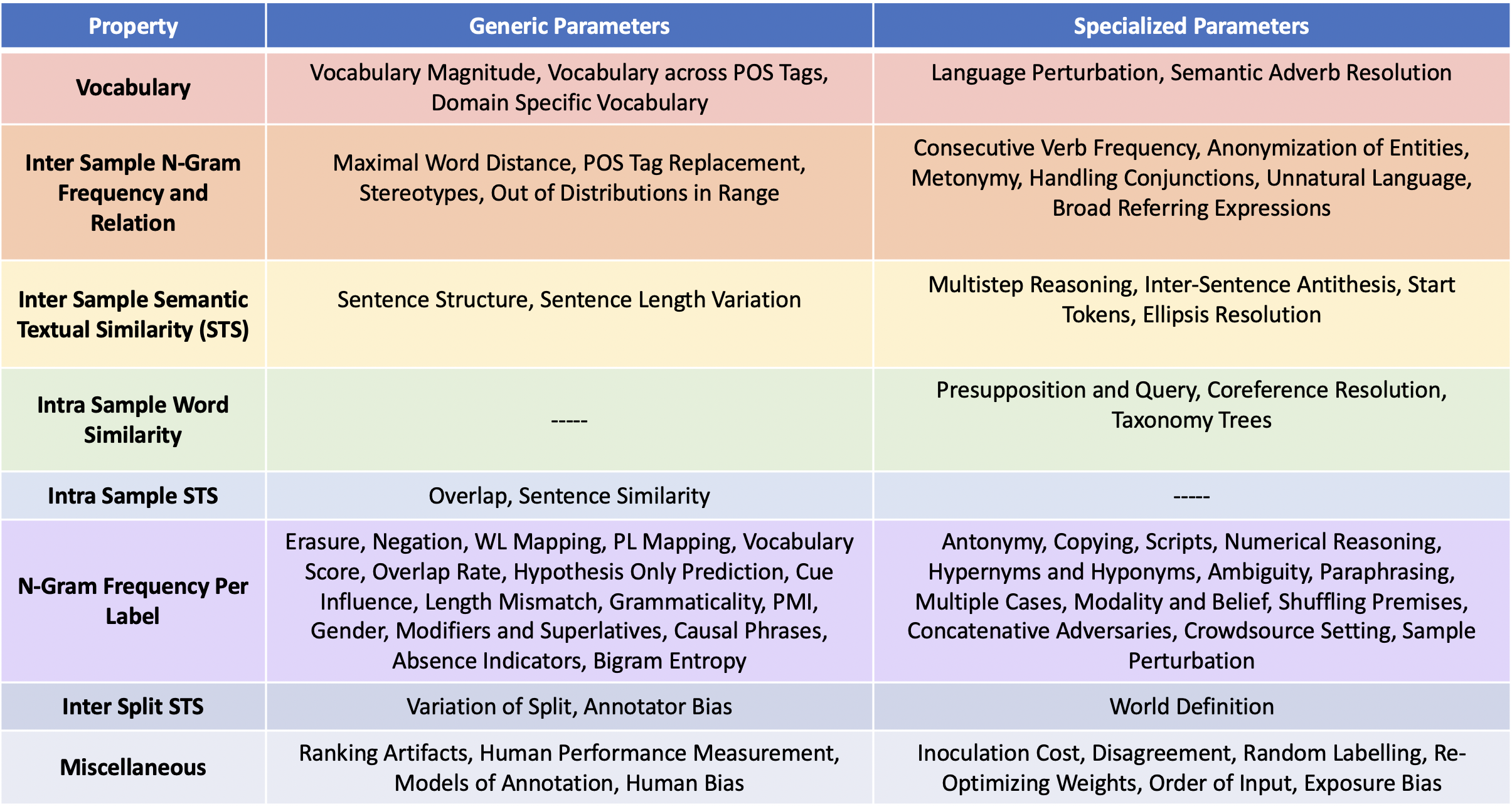}
  \caption{Parameter categorization: Generic parameters are task and model independent unlike specialized ones.}
  \label{fig:three}
% \vspace{-2mm}
\end{figure*}
\paragraph{Human Bias}
In a popular de-biasing approach, prior knowledge is used to develop a naive model which is trained to exploit biases. Subsequently, this model is combined with a robust model and the ensemble model is trained \cite{clark2019don, he2019unlearn, mahabadi2019simple}. However, this process might induce bias due to the involvement of humans and their knowledge. 
%as it requires prior knowledge of humans.
%because of the involvement of humans in utilizing 
% Training a naive model based on 
% They use a prior knowledge of biases to train a naive model that exploits
% dataset biases. Then this model is combined with a robust model, and the ensemble is trained \cite{clark2019don}.
\section{Discussion: Towards a Generic Metric}
% We categorize potential leads into three parts (i) Generic Leads (ii)
Certain parameters are specific to models or tasks. We divide parameters into two categories: (i) generic, and (ii) specialized, as shown in Figure \ref{fig:three}. 
%Those might help in probing models and analyzing bias better, and thus can be used as guidelines in creating bias-minimized data or tools to visualize the bias exploitation process in models. However, these can not be included while developing a generic metric. Figure \ref{fig:three} enlists both generic and specialized parameters.
% have to be updated every time we have a new SOTA model. So, we don't include them in our development of generic DQI. Table \ref{lcategory1} and \ref{lcategory2} enlists filtered parameters across categories. We use parameters to extend our intuition, but don't rely on them completely. For example, we don't consider any parameters for the Intra-sample Word Similarity Category.
\vspace{-2mm}
%\paragraph{Hyper-parameters and Genericness of the Metric:}
\paragraph{The Generic Metric:}
The generic metric is the desired and reliable solution to evaluate diverse benchmarks.
The effect of generic parameters, identified using this survey, can be converted to a mathematical formula, resulting in an empirical metric. The metric will have hyper-parameters in the form of boundaries separating high and low quality data (i.e., inductive and spurious bias). As boundaries are dependent on applications, BiomedicaNLP for example, might have lower tolerance levels for spurious bias than General NLP. This is similar to quality indices in domains such as water where the quality of water needed for irrigation is different than that of drinking or medicine. 
% In concurrent works, Anonymous~[1] follows a similar procedure to implement a generic metric, which is subsequently used by Anonymous~[2, 3] to develop paradigms for the next generation of benchmarks.

% which needs to be tuned in a supervised setting for few samples. 

% In various other domains such as water, food, and power, we do have hyper-parameters in the quality indices. This is because of the dependence of a quality index on its application; for example, in the case of water quality, the quality of water needed for irrigation is different from the quality of water used for drinking, skin care, fitness and medicine. Thus, the allowed limits of water components varies according to the use case. Similarly, we should have hyper-parameters in our metric, determining the tolerance of its components. These must be tuned for different NLP tasks and domains; for example, hyper-parameters for Biomedical NLP may be very different from those used for general NLP. 
% However, the metric is still generic as the formula does not change across task or dataset.
%\footnote{SOTA Model such as ROBERTA} 
\vspace{-2mm}
%\paragraph{Opportunity for Model-specific Metric using Active Learning:}
\paragraph{Specialized Metric:} 
Model or task specific metrics can be built from the specialized parameters, following the same process as the generic metric. Application of this metric in an active learning setup, where the error analysis of models can be used to tune hyper-parameters, enables control over the hardness of benchmarks without overloading humans/ risking the possibility of human bias, unlike existing works \cite{nie2019adversarial}. The metric can also help in probing models, and developing the next generation of language models that can better utilize inductive and spurious bias.

% We include model-specific parameters since they can be utilized as constraints to develop model-specific metric which can be further utilized in creating hard datasets or understanding bias in models. The idea is there should be a minimum number of patterns which are difficult for the SOTA model to crack while solving a dataset. For example, Semantic Adverbs should be present a minimum number of times in a dataset. 
% %Those might help in probing models and analyzing bias better, and thus can be used as guidelines in creating bias-minimized data or tools to visualize the bias exploitation process in models. However, these can not be included while developing a generic metric.
% % The same is true for Domain Specific Words as they force models to learn and not look for patterns. Similarly, Consecutive Verb Frequency should have a minimum threshold for certain verbs. Also there should be sufficient number of figures of speech.  This is to force models to not rely on spurious biases in order to solve that dataset.
% Active Learning can be used to make the dataset harder using errors that a model make to retune hyper-parameters in our metric. Active learning can help to partially automate the feedback process, reduce the load on crowd workers and minimize human bias. 
\section{Conclusion} 
We survey recent works on bias to set the stage for the development of data quality metrics that can be used to evaluate benchmarks. We identify a set of seven language properties which can represent various possible sample interactions (and therefore biases) in a benchmark. Corresponding to the language properties, we identify and curate 68 potential parameters that capture bias properties, including origins, types, impact on performance, and generalization across diverse NLP tasks. We further categorize these parameters into (i) generic parameters which can be integrated to a generic metric, and (ii) parameters which can be leveraged for a model/task/dataset-specific metric. Our survey identifies gaps in the current research and suggests that future research directed towards concepts such as a metric quantifying quality could open up opportunities for the next generation of benchmarks, models, and evaluation metrics.

\section{Limitation}
We propose parameters that can make a data quality metric, however the feasibility of such a metric is unknown, even though we have quality metrics in other domains such as water, food, air etc. However, we believe the parameters will still be helpful to look for hidden biases in data.

In our survey, we try to cover most of the papers related to spurious bias and data quality, but it is possible that we might have missed some papers.

Spurious bias is prevalent across languages and is probably more likely in low-resource language because of lack of benchmarks, guidelines, evaluation etc. This survey is limited to english benchmarks, probably because most of the existing studies on spurious bias is performed on english benchmarks. Future work should involve other languages, specifically the low resource languages as part of the analysis.

% \footnote{Detailed Analysis is in Supplemental materials}
% Entries for the entire Anthology, followed by custom entries
\bibliography{anthology,custom}
\bibliographystyle{acl_natbib}

\appendix

\section{Supplemental Material}
% \subsection{Anonymous et. al.~[1]}
\subsection{Justification of Task Ordering}
NLI takes a two sentence input (premise and hypothesis), to output a single label (entailment, neutral, and contradiction). Argumentation takes a four sentence input - claim, reason, warrant, and alternative warrant- and outputs the choice between the warrant and the alternative warrant. Multiple choice questions read either single/multi-line inputs and a set of choices comprising of words/sentences; they output a single choice (number/word/sentence). Open ended questions always output words/one or more sentences, after reading a multi-line input. Reading comprehension questions follow the same patterns as regular question answering samples, in that a multi-line input is read, and a choice/word/phrase/sentence is the output. The output format depends on the patterns of questions asked such as fill in the blanks and sentence completion. Also, the volume of input read is generally much larger than that seen in question answering.  Finally, abstractive summarization deals with both multi-line input and multi-line output.
\label{sec:supplemental}
\subsection{Illustrating Parameters with Examples}
Here, we provide more details of each of the 68 parameters along with examples for better illustration.
\label{sec:eg1}\paragraph{Vocabulary Magnitude:}
A dataset of size of 100k samples and 30k unique words will have a vocabulary magnitude of 0.3.
\label{sec:eg5}\paragraph{Language Perturbation:}
The substitution of words like 'and' or 'by' with fillers such as 'blah' helps check if the original words are being used as a part of the reasoning context or not.
\label{sec:eg6}\paragraph{Semantic Adverb Resolution:}
There is a difference in the contexts created by 'always', 'sometimes', 'often', and 'never.'
\label{sec:eg7}\paragraph{Domain Specific Vocabulary:}
The names of countries such as Syria, Canada, Mexico, etc., and nationalities, such as Indian, Swiss, etc. are not recognized by language models, and performance on instances containing these words is low.
\label{sec:eg2}\paragraph{Maximal Word Distance:}
A dataset that covers the scientific domain will have words dissimilar to more commonly used language.
\label{sec:eg3}\paragraph{POS Tag Replacement:}
Consider the word 'Turkey' in vocabulary, where the context is that Turkey refers to the country. An equivalent country name (of the same POS tag) like 'Russia' can be used for replacement. Turkey could also refer to a bird, such as 'Turkey dinner'. In this case, on replacement, 'Russia dinner' will be generated. This case does not add an example that makes sense. So such samples are discarded based on the count of the bigrams generated on replacement. In TextFooler, consider the input “The solution while  viable was totally over the budget.The solution while  viable was totally over the budget.” The output might be: “The arrangement whereas reasonable was completely over the budget.”
\label{sec:eg4}\paragraph{Consecutive Verb Frequency:}
It has been observed that on translation from English to German and back, sentences such as 'He was running stressed for work' drop the second verb on retranslation, and becoming 'He was running for work.'
\label{sec:eg8}\paragraph{Anonymization of Entities:}

Original Version: Content: 'The XYZ director allegedly struck by Brock Paine will not press charges against the “Guess Who” host.' Question: Who hosts Guess Who? Answer:Brock Paine

Anonymized Version: Content: 'The $ent1$ producer allegedly struck by $ent2$ will not press charges against the “Guess Who” host.' Question: Who hosts Guess Who? Answer: $ent2$
\label{sec:eg12}\paragraph{Metonymy:}
'Let me give you a hand.' Here, hand refers to help.
\label{sec:eg18}\paragraph{Stereotypes:}
Word associations like 'cook' or 'dolls' with 'girls', or 'temples' with 'India' are a source of bias.
\label{sec:eg21}\paragraph{Out of Distributions in Range:}
'Sheila and I' and 'Sheila or I' have different contextual meanings which can't be solved by pattern correlation. 
'Jim, John and Bob are 14, 12, and 18. Who is the second oldest?' returns the correct answer. But if their ages are '1997', '2001', and '2010', then the system returns the wrong answer.
\label{sec:eg22}\paragraph{Unnatural Language:}
The sentence: 'She was [MASK] rich, she was also a diva,' has different meanings if you substitute 'not' or 'very' in it.
\label{sec:eg49}\paragraph{Broad Referring Expressions:}
Generic terms like 'this', 'the', 'that', or 'it'  can be used to refer to objects on different occasions. These must be resolved to remove ambiguity.
\label{sec:eg9}\paragraph{Sentence Structure:}
If a majority of sentence structures follow passive voice, an active voice sentence won't be easily parsed.
\label{sec:eg10}\paragraph{Multistep Reasoning:}
'When comparing a 23, a 38 and a 31 year old, the [MASK] is oldest A. second B. first C. third.'
\label{sec:eg11}\paragraph{Inter-Sentence Antithesis:}
'It was [MASK] nice, it was really inconvenient. A. not B. really.'
\label{sec:eg14}\paragraph{Sentence Length Variation:}
Sentences with less detail are shorter, and therefore more likely to be classified as entailment.
\label{sec:eg15}\paragraph{Start Tokens:}
The candidate answer resolution is restricted by starting “wh-” and “how many" expressions.
\label{sec:eg16}\paragraph{Ellipsis Resolution:}
'I’ll order the linguini and you can too.' can be unrolled as 'I’ll order the linguini and you can order the linguini too.'
\label{sec:eg13}\paragraph{Presupposition and Query:}
'This ban is the first ban for the sale of fur of endangered species in Canada.' Here, the statement assumes that there is a ban, and the model must reason on whether the ban was the first, not on the existence of the ban.
\label{sec:eg17}\paragraph{Coreference Resolution:}
'Tom said that he would get it done.'Here, he refers to Tom.
\label{sec:eg19}\paragraph{Taxonomy Trees:}
'Horse and crow' are grouped as animal, but 'crow and horse' are grouped as birds. This is because 'crow' is closer to 'bird' on the taxonomy tree than 'animal.'
\label{sec:eg20}\paragraph{Overlap:}
'The dog sat on the mat' and 'The dog did not sit on the chair' contain significant overlap and hence can easily be solved. In HANS, consider the premise 'The teachers heard the students left' and 'The teachers heard the students'. If a model relied on overlap, it would mark this sample as entailment, even though the gold label is neutral. 
\label{sec:eg55}\paragraph{Erasure:}
Consider the sample 'I took my daughter and her step sister to see a show at Webster hall . It is so overpriced I’m in awe.' Using a BI-LSTM, the minimal set of words identified for 'value' is 'It is so overpriced I’m in awe.'
\label{sec:eg24}\paragraph{Similarity:}
Similarity indicates overlapping detail. For example, 'The bird sang' and 'The robin warbled outside the window as it looked for breakfast' have less overlap due to the presence of more detail in the second sentence.
\label{sec:eg25}\paragraph{Negation:}
The sentences 'She was pleased' and 'She could do nothing that did not please her' might be labeled as contradiction 
due to the presence of negation terms.
\label{sec:eg26}\paragraph{Antonymy:}
Simple binary opposites are 'hot' and 'cold'. Less direct opposites are words like 'winter' and 'summer'.
\label{sec:eg27}\paragraph{WL Mapping:}
'Humans' and' instruments are found to be indicators of entailment, 'tall' and 'win' that of neutral, and 'sleep' and 'no' of contradiction.
%\begin{equation}
$P(l/w)=\frac{p(w,l)}{p(w)\cdot p(l)}$
%\end{equation}

\label{sec:eg28}\paragraph{PL Mapping:}
For the phrase ‘x was sentient….’ ; by identifying the nature of ‘x’, a model can infer the label without looking at the rest of the sentence . Such lexical semantic exploitation indicates that context is not used in solving such samples.
%\begin{equation}
$P(l/p)=\frac{p(p,l)}{p(p)\cdot p(l)}$
%\end{equation}

\label{sec:eg29}\paragraph{Vocabulary Score:}
Consider the word 'move' in the entailment, neutral, and contradiction classes, with counts 200, 345, and 126 respectively. Then, the score vector would be [3 200 345 126].

\label{sec:eg31}\paragraph{Copying:}
Copy all possible subset of words from the premise to the hypothesis iteratively, and check when the label changes.
\label{sec:eg32}\paragraph{Hypothesis Only Prediction:}
The sample: 'People raise dogs because they are obedient' and 'People raise dogs because dogs are obedient', benefits from considering hypothesis only as there is no coreference to be resolved.
\label{sec:eg33}\paragraph{Cue Influence:}
Let $k$ be a cue, $T_{j}$ be the set of tokens in the warrant for data point $i$  with label $j$, and $n$ be the total number of data samples.

Applicability: number of data points a cue occurs with one label but not the other
%\begin{equation}
$\alpha _{k}= \sum_{i=1}^{n} [\exists j,k \in T_{j}^{(i)}  k \notin T_{\neg j}^{(i)}]$
%\end{equation}
Productivity: proportion of applicable data points for which a cue predicts the correct answer
%\begin{equation}
$\pi _{k}= \frac{\sum_{i=1}^{n} 1[\exists j,k \in T_{j}^{(i)} \wedge k \notin T_{\neg j}^{(i)} \wedge y_{i}=j]}{\alpha_{k}}$
%\end{equation}
Coverage: proportion of applicable cases of a cue over the total number of data points
%\begin{equation}
$\xi _{k}= \frac {\alpha_{k}}{n}$
%\end{equation}

\label{sec:eg34}\paragraph{Length Mismatch:}
The sample: 'She was happy with her bonus’ and 'She decided to celebrate her raise at work by eating out,’ is more likely to be labelled as neutral.
\label{sec:eg35}\paragraph{Grammaticality:}
Consider the sample’: She has no option’ and She has no way than the others’. This is more likely to be classified as 'non-entailment.'
\label{sec:eg36}\paragraph{PMI:}
%\begin{equation}
$PMI(word,label)=\log\frac{p(word,label)}{p(word)\cdot p(label)}$
%\end{equation}

\label{sec:eg37}\paragraph{Scripts:}
Consider the sample: 'Canada's plans to launch a satellite, but U.S. officials say the launch is a disguised long-range missile test' and 'The U.S. fears that the Canadian satellite is a ruse to hide the testing of a missile.' There is a familiar script at play here. Countries want to test  military equipment, but don't want to be seen as testing them, so may try and hide or cover up the test. Other countries are worried about this form of deceit, and may try and put political pressure on the testing country in order to prevent deceit.
\label{sec:eg38}\paragraph{Numerical Reasoning:}
'There were two major concerts last week in Tulsa, with 10000 people attending the first one in Oakland Center and more than 8000 at The Gathering Place.' Requires a sum of 30+10 to be calculated to address the hypothesis: 'Last week there were 2 concerts in Tulsa, attended by more than 18000 people.'
\label{sec:eg39}\paragraph{Gender:}
Using terms like 'woman' and 'boy' instead of 'person' or 'child' are indicative of non-entailment.
\label{sec:eg40}\paragraph{Hypernyms and Hyponyms:}
(i) Words like 'wolf' and 'dog' are both animals, but confusion may occur during hyponym resolution as a wolf is a wild animal.
(ii) A table might serve as a superset for its legs, which is not a true hypernym.
\label{sec:eg41}\paragraph{Modifiers and Superlatives:}
Words like 'tall' or 'popular' and 'best' or 'first' are indicative of neutral label.
\label{sec:eg42}\paragraph{Causal Phrases:}
Sentences that contain causal words like 'due to', 'because of', 'consequently', etc. are indicative of neutral label.
\label{sec:eg43}\paragraph{Absence Indicators:}
The word 'sleep' indicates the absence of activity, and hence is used as an indicator of contradiction.
\label{sec:eg44}\paragraph{Ambiguity:}
'She had a black bat' requires context and knowledge to decide if 'bat' refers to an animal, or sports equipment.
\label{sec:eg45}\paragraph{Bigram Entropy:}
% \begin{equation}
% $H(c | w) = −\sum_{c}^{}p(c | w)\log p(c | w)    $
% \end{equation}
Object bias: For example, 'playing cello' is the only class depicting cellos. This can be inferred by searching for 'cello' or 'music'. 

Scene bias: For example, 'soccer juggling' can be resolved by searching for words like 'goal', 'net', or 'ball'. 

Person bias: For example, 'military marching' can be resolved by matching to words like 'army' or 'parade'.

\label{sec:eg46}\paragraph{Paraphrasing:}
'Same' and 'replica' are paraphrases, but 'same' and 'about same' are not.

PAWS: Word Swapping: 'Can a bad person have good habits? : 'Can a good person have bad habits?'

PAWS: Back Translation: 'The band also toured in Asia in 1983.' : 'In 1983, the band also toured in Asia.'

\label{sec:eg47}\paragraph{Multiple Cases:}

Context: [...] This order is for Table Number 93. [...] Question: What was the number of the table for which the order was taken?
Answer: 93
Possible answers: Table 93, Table Number 93, Number 93, 93
Here, multiple choices have the correct span of 93 \cite{trischler-etal-2017-newsqa}.
\label{sec:eg48}\paragraph{Modality and Belief:}

Epistemic: Agatha must be the murderer. (necessity:neutral)

Deontic: Agatha must go to jail. (obligatory:neutral)

Circumstantial: Agatha must sneeze. (possibility:entailment)

Belief for the above case is true/false in order to label them.
\label{sec:eg51}\paragraph{Shuffling Premises:}
It is a method of iteratively substituting premises to check word correlation.
\label{sec:eg52}\paragraph{Concatenative Adversaries:}
Add distractor words at the end of hypotheses such as negation, superlatives, etc. to test the model's operation over the original samples.
\label{sec:eg56}\paragraph{Crowdsource Setting:}
The length of a contradiction hypothesis is generally shorter than that of the original premise, and it uses simpler language. 

\label{sec:eg57}\paragraph{Sample Perturbation:}

Counterfactual Sample:

P: A woman crouches on the lake shore while filling a pot.

OH: A woman fills a pot in the lake while camping. (Neutral)

NH: A woman fills a pot in the lake. (Entailment)

Contrast Set Sample:

Original Text: Two similarly-colored and similarly-perched parrots are on a tree branch.

New Text: Two differently-colored but similarly-perched parrots are on a tree branch.

\label{sec:eg53}\paragraph{Variation of Split:}
Different split variations are required for proper benchmarking, to ensure a true accuracy increase.
%\begin{equation}
$\widehat{\delta }=\mathit{M(G_{test},S_{1})} - \mathit{M(G_{test},S_{2})}$
%\end{equation}

Accuracy difference: $\widehat{\delta}$

Model: $M$

Test Set: $G_{test}$ 

Systems 1 and 2: $S_{1},S_{2}$

\label{sec:eg54}\paragraph{Inoculation Cost:}

Adversarial NLI:

Premise:A melee weapon is any weapon used in direct hand-to-hand combat; by contrast with ranged weapons which act at a distance. The term “melee” originates in the 1640s from the French word, which refers to hand-to-hand combat, a close quarters battle, a brawl, a con- fused fight, etc. Melee weapons can be broadly divided into three categories

Hypothesis: Melee weapons are good for ranged and hand-to-hand combat.

\label{sec:eg61}\paragraph{Disagreement:}
A particular annotator overuses the label of entailment, and marks very few samples as neutral. This pattern can be used as a bias by a model.

% % \begin{figure*}
% % \includegraphics[width=\linewidth,height=20cm]{lt1.png}
% %   \caption{Detailed Information on Parameters - I}
% % \label{fig:LeadTable1}
% % \end{figure*}

% % \pagebreak

% % \begin{figure*}
% % \includegraphics[width=\linewidth,height=20cm]{lt2.png}
% %   \caption{Detailed Information on Parameters - II}
% % \label{fig:LeadTable2}
% % \end{figure*}

% % \pagebreak

% % \begin{figure*}
% % \includegraphics[width=\linewidth,height=15cm]{lt3.png}
% %   \caption{Detailed Information on Parameters - III}
% % \label{fig:LeadTable3}
% % \end{figure*}

\clearpage
\onecolumn
\begin{small}
\centering
\begin{longtable}{|p{0.6in}|p{0.75in}|p{0.5in}|p{0.5in}|p{0.5in}|p{1.5in}|p{0.55in}|}
\label{tab:long} 
\\
\hline \multicolumn{1}{|c|}{\textbf{Category}} & \multicolumn{1}{c|}{\textbf{Parameter}} & 
\multicolumn{1}{c|}{\textbf{Desired Level}} & \multicolumn{1}{c|}{\textbf{Dataset}} & \multicolumn{1}{c|}{\textbf{Highest Task}} & 
\multicolumn{1}{c|}{\textbf{Source}} & 
\multicolumn{1}{c|}{\textbf{Type}} \\ \hline 
\endfirsthead

\multicolumn{7}{c}%
{{\bfseries \tablename\ \thetable{} -- continued from previous page}} \\
\hline \multicolumn{1}{|c|}{\textbf{Category}} & \multicolumn{1}{c|}{\textbf{Parameter}} &
\multicolumn{1}{c|}{\textbf{Desired Level}} & 
\multicolumn{1}{c|}{\textbf{Dataset}} & 
\multicolumn{1}{c|}{\textbf{Highest Task}} & 
\multicolumn{1}{c|}{\textbf{Source}} & 
\multicolumn{1}{c|}{\textbf{Type}} \\ \hline 
\endhead

\hline \multicolumn{7}{|r|}{{Continued on next page}} \\ \hline
\endfoot

\hline \hline
\endlastfoot
Vocabulary
 & Vocabulary Magnitude &
  High &
  SNLI, MNLI &
  Summarization &
  Created &
  Measure \\ \cline{2-7} 
 & Vocabulary Across POS Tags &
  High &
  - &
  Summarization &
  Created &
  Measure \\ \cline{2-7} 
 &
  Language Perturbation &
  Low Response &
  RoBERTa Pretrained Corpus&
  Summarization &
  Modified from \cite{talmor2019olmpics} &
  Technique \\ \cline{2-7} 
&
  Semantic Adverb Resolution &
  High Response &
  RoBERTa Pretrained Corpus &
  Summarization &
  Compiled from \cite{talmor2019olmpics} &
  Feature \\ \cline{2-7} 
&
  Domain Specific Vocabulary &
  High &
  SNLI &
  Summarization &
  Created &
  Feature \\ \hline
%  ---------------------------------------------------------------------------------------------- 
  Inter-sample N-gram Frequency and Relation
& Maximal Word Distance &
  High &
  SNLI, MNLI &
  Summarization &
  Created &
  Measure  \\ \cline{2-7}
 &
  POS Tag Replacement &
  High &
  Twitter, SNLI &
  Summarization &
  Compiled from \cite{ribeiro-etal-2018-local,zhao2017generating,glockner2018breaking,jin2019bert}&
  Technique \\ \cline{2-7}   
&
  Consecutive Verb Frequency &
  High &
  Google Translate &
  Summarization &
  Compiled from \cite{zhao2017generating} &
  Feature \\ \cline{2-7} 
  &
  Anonymization of Entities &
  High Response &
  Daily Mail &
  Summarization &
  Modified from \cite{hermann2015teaching,li2018resound} &
  Feature \\ \cline{2-7}
 &
  Metonymy &
  High Response &
  RTE-5 &
  Summarization &
  Modified from \cite{clark2018knowledge} &
  Feature \\ \cline{2-7} 
 &
  Stereotypes &
  Low Response &
  SNLI &
  Summarization &
  Modified from \cite{rudinger-etal-2017-social} &
  Feature \\ \cline{2-7} 
 &
  Out of Distributions in Range &
  Low &
  SNLI &
  Summarization &
  Compiled from \cite{talmor2019olmpics} &
  Feature \\ \cline{2-7} 
 &
 Handling Conjunctions &
 High Response &
 SNLI &
 Summarization &
 Compiled from \cite{talmor2019olmpics} &
 Feature \\ \cline{2-7}
 &
  Unnatural Language &
  High Response &
  SNLI &
  Summarization &
  Compiled from \cite{talmor2019olmpics} &
  Feature \\ \cline{2-7}
 &
  Broad Referring Expressions &
  High Response &
  - &
  Summarization &
  Compiled from \cite{degen2020redundancy} &
  Feature \\ \hline
%  ---------------------------------------------------------------------------------------------- 
Inter-sample STS
&  Sentence Structure &
  Uniform Distribution &
  SNLI &
  Summarization &
  Compiled from \cite{poliak2018hypothesis} &
  Feature \\ \cline{2-7} 
 &
  Multistep Reasoning &
  High Response &
  SNLI &
  Summarization &
  Compiled from \cite{talmor2019olmpics,naik2018stress} &
  Feature \\ \cline{2-7} 
 &
  Inter-Sentence Antithesis &
  High Response &
  SNLI &
  Summarization &
  Compiled from \cite{naik2018stress} &
  Feature \\ \cline{2-7} 
 &
  Sentence Length Variation &
  Diverse &
  SNLI &
  Summarization &
  Compiled from \cite{gururangan2018annotation} &
  Feature \\ \cline{2-7} 
  &
  Start Tokens &
  Diverse &
  MARCO &
  Summarization &
  Modified from \cite{sugawara2018makes} &
  Feature \\ \cline{2-7}  
  &
  Ellipsis Resolution &
  High Response &
  SNLI &
  Summarization &
  Compiled from \cite{clark2018knowledge} &
  Feature \\ \hline 
%  ---------------------------------------------------------------------------------------------- 
Intra-sample Word Similarity
 &
  Presupposition and Query &
  High Response &
  RTE-5 &
  Summarization &
  Modified from \cite{clark2018knowledge} &
  Feature \\ \cline{2-7} 
 &
  Coreference Resolution &
  High Response &
  SNLI &
  Summarization &
  Compiled from \cite{gururangan2018annotation,cirik2018visual} &
  Feature \\ \cline{2-7} 
  &
  Taxonomy Trees &
  Low Response &
  SNLI &
  Summarization &
  Modified from \cite{talmor2019olmpics} &
  Feature \\ \hline
 %  ---------------------------------------------------------------------------------------------- 
 Intra-Sample STS
 &
  Overlap &
  Low &
  SNLI &
  Summarization &
  Compiled from \cite{naik2018stress,mccoy2019right} &
  Measure \\\cline{2-7}
  &
  Sentence Similarity &
  Low &
  SNLI &
  Summarization &
  Compiled from \cite{naik2018stress,clark2018knowledge} &
  Measure \\  
  \hline
 %  ---------------------------------------------------------------------------------------------- 
  N-gram Frequency per Label
& Erasure &
High Response &
Stanford Sentiment Treebank &
Summarization &
Modified from \cite{li2016understanding} &
Technique \\ \cline{2-7} 
 &
  Negation &
  Low &
  SNLI &
  Summarization &
  Compiled from \cite{poliak2018hypothesis} &
  Feature \\ \cline{2-7} 
 &
  Antonymy &
  High Response &
  SNLI &
  Summarization &
  Compiled from \cite{naik2018stress} &
  Feature \\ \cline{2-7} 
 &
  WL Mapping &
  Low &
  SNLI &
  Summarization &
  Compiled from \cite{poliak2018hypothesis} &
  Measure \\ \cline{2-7} 
 &
  PL Mapping &
  Low &
  SNLI &
  Summarization &
  Compiled from \cite{poliak2018hypothesis} &
  Measure \\ \cline{2-7} 
 &
  Vocabulary Score &
  High &
  SNLI &
  Argumentation &
  Created &
  Measure \\ \cline{2-7}   
 &
  Copying &
  High Response &
  Wall Street Journal- Penn Treebank &
  Summarization &
  Modified from \cite{gu-etal-2016-incorporating,see-etal-2017-get,merity2016pointer,aharoni-goldberg-2018-split} &
  Technique \\ \cline{2-7} 
 &
  Hypothesis Only Prediction &
  Low Response &
  SNLI &
  NLI &
  Compiled from \cite{tan2019investigating} &
  Technique \\ \cline{2-7} 
 &
  Cue Influence &
  Low Response &
  ARCT &
  Argumentation &
  Compiled from \cite{niven2019probing} &
  Measure \\ \cline{2-7} 
 &
  Length Mismatch &
  Diverse &
  SNLI &
  Summarization &
  Compiled from \cite{poliak2018hypothesis,gururangan2018annotation,naik-etal-2018-stress} &
  Feature \\ \cline{2-7} 
 &
  Grammaticality &
  High &
  FN+ &
  Summarization &
  Compiled from \cite{poliak2018hypothesis} &
  Feature \\ \cline{2-7} 
 &
  PMI &
  Low &
  SNLI &
  Summarization &
  Compiled from \cite{naik2019exploring,gururangan2018annotation} &
  Measure \\ \cline{2-7} 
 &
  Scripts &
  High Response &
  RTE-5 &
  Summarization &
  Compiled from \cite{clark2018knowledge} &
  Technique \\ \cline{2-7} 
 &
  Numerical Reasoning &
  Uniform Distribution &
  SNLI &
  Summarization &
  Compiled from \cite{naik2018stress,gururangan2018annotation} &
  Feature \\ \cline{2-7} 
 &
  Gender &
  Uniform Distribution &
  SNLI &
  Summarization &
  Compiled from \cite{gururangan2018annotation} &
  Feature \\ \cline{2-7} 
 &
  Hypernyms and Hyponyms &
  High Response &
  SNLI &
  Summarization &
  Modified from \cite{glockner2018breaking,richardson2019does,levy2015supervised} &
  Feature \\ \cline{2-7} 
 &
  Modifiers and Superlatives &
  Uniform Distribution &
  SNLI &
  Summarization &
  Compiled from \cite{gururangan2018annotation} &
  Feature \\ \cline{2-7} 
 &
  Causal Phrases &
  Uniform Distribution &
  SNLI &
  Summarization &
  Compiled from \cite{gururangan2018annotation} &
  Feature \\ \cline{2-7} 
 &
  Absence Indicators &
  Uniform Distribution &
  SNLI &
  Summarization &
  Compiled from \cite{gururangan2018annotation} &
  Feature \\ \cline{2-7} 
 &
  Ambiguity &
  High Response &
  SNLI &
  Summarization &
  Compiled from \cite{naik2018stress} &
  Feature \\ \cline{2-7} 
 &
  Bigram Entropy &
  Low &
  SNLI &
  Summarization &
  Compiled from \cite{tan2019investigating,li2018resound} &
  Measure \\ \cline{2-7} 
 &
  Paraphrasing &
  High Response &
  SNLI &
  Summarization &
  Compiled from \cite{clark2018knowledge,sugawara2018makes,zhang2019paws} &
  Technique \\ \cline{2-7} 
 &
  Multiple Cases &
  High Response &
  NewsQA &
  Summarization &
  Modified from \cite{sugawara2018makes,trischler-etal-2017-newsqa} &
  Technique \\ \cline{2-7} 
 &
  Modality and Belief &
  High Response &
  SNLI &
  Summarization &
  Created &
  Feature \\ \cline{2-7}
  &
  Shuffling Premises &
  High Response &
  SNLI &
  Summarization &
  Compiled from \cite{tan2019investigating} &
  Technique \\ \cline{2-7} 
 &
  Concatenative Adversaries &
  High Response &
  SNLI &
  Summarization &
  Compiled from \cite{naik2018stress,jia2017adversarial} &
  Technique \\ \cline{2-7}
  &
  Crowdsource Setting &
  Low Response &
  Story Cloze &
  Summarization &
  Compiled from \cite{schwartz-etal-2017-story,qin2004exploratory,yancheva2013automatic,newman2003lying} &
  Technique \\ \cline{2-7} &
 Sample Perturbation &
  High Response &
  DROP, IMDb, SNLI &
  Summarization &
  Compiled from \cite{naik2018stress,jia2017adversarial} &
  Technique \\ \hline
%  ---------------------------------------------------------------------------------------------- 
Inter-split STS
  &
  Variation of Split &
  High Response &
  SNLI &
  Summarization &
  Compiled from \cite{tan2019investigating,gorman-bedrick-2019-need} &
  Technique \\ \cline{2-7}
  &
  Annotator Bias &
  High Response &
  MNLI, OpenBookQA, CommonsenseQA &
  Summarization &
  Compiled from \cite{geva2019we} &
  Technique \\ \cline{2-7}
  &
  World  Definition &
  High Response &
  PASCAL 07 &
  Summarization &
  Compiled from \cite{torralba2011unbiased} &
  Technique \\ \hline
%  ---------------------------------------------------------------------------------------------- 
  Miscellaneous
  &
  Innoculation Cost &
  High Response &
  Open Domain QA, Adversarial NLI &
  Summarization &
  Compiled from \cite{richardson2019does,nie2019adversarial} &
  Technique \\ \cline{2-7}  
&
  Disagreement &
  Low Correlation &
  - &
  Summarization &
  Compiled from \cite{reidsma2008reliability} &
  Feature \\ \cline{2-7}  
  &
  Random Labelling &
  Low Response &
  CIFAR10, ImageNet &
  Summarization &
  Compiled from \cite{zhang2016understanding} &
  Technique \\ \cline{2-7}  
  &
  Re-Optimizing Weights &
  High Response &
  Colored MNIST, Kinetics, &
  Summarization &
  Compiled from \cite{li2019repair} &
  Technique \\ \cline{2-7}  
  &
  Adversarial Filtering &
  Low Response &
  SNLI, MNLI &
  Summarization &
  Compiled from \cite{bras2020adversarial} &
  Technique \\ \cline{2-7}  
  &
  Ranking Artifacts &
  - &
  - &
  Summarization &
  Created &
  Technique \\ \cline{2-7}  
  &
  Human Performance Measurement &
  - &
  Contrast Sets &
  Summarization &
  Created &
  Technique \\ \cline{2-7}  
  &
  Order of Input &
  - &
  GLUE &
  Summarization &
  Compiled from \cite{dodge2020fine} &
  Technique \\ \cline{2-7} 
  &
  Models of Annotation &
  - &
  - &
  Summarization &
  Compiled from \cite{paun2018comparing} &
  Technique \\ \cline{2-7} 
  &
  Exposure Bias &
  - &
  - &
  Summarization &
  Compiled from \cite{caccia2018language} &
  Technique \\  \cline{2-7} 
    &
  Human Bias &
  Low Response &
  MNLI, VQA, HANS &
  Summarization &
  Compiled from \cite{clark2019don,he2019unlearn,mahabadi2019simple} &
  Technique \\  
%   \\
  \hline  

\caption{Detailed Information on Parameters} 
\end{longtable}
\end{small}

% \end{document}

% \section{Example Appendix}
\label{sec:appendix}

% This is an appendix.

\end{document}